\documentclass[10pt,twocolumn,letterpaper]{article}

\usepackage{cvpr}              

\usepackage[table]{xcolor}

\definecolor{cvprblue}{rgb}{0.21,0.49,0.74}

\newcommand{\minisection}[1]{\vspace{1mm} \noindent \textbf{#1} \hspace{0mm}}
\def\paperID{6603} 
\def\confName{CVPR}
\def\confYear{2024}

\usepackage{adjustbox}     
\usepackage{multirow}
\definecolor{mygray}{gray}{0.95}
\usepackage{hhline}
\usepackage{nicematrix}
\usepackage{xr-hyper}
\usepackage{amssymb}
\usepackage{pifont}
\usepackage[pagebackref,breaklinks,colorlinks,citecolor=cvprblue]{hyperref}
\newcommand{\methodName}{ProMark\xspace}
\newcommand{\vect}[1]{\boldsymbol{#1}}
\newcommand{\cmark}{\checkmark}
\newcommand{\xmark}{\ding{53}}

\title{ProMark: Proactive Diffusion Watermarking for Causal Attribution}

\author{Vishal Asnani$^{1,2}$ \\
\and John Collomosse$^{1,3}$\\
\and Tu Bui$^{3}$\\
\and Xiaoming Liu$^{2}$\\
\and Shruti Agarwal$^{1}$ \\
\and \vspace{-1.1cm}\\
$^{1}$Adobe Research, $^{2}$Michigan State University,
$^{3}$University of Surrey\\
{\tt\small \ \{asnanivi,liuxm\}@msu.edu~~ \{collomos,shragarw\}@adobe.com \ ~t.v.bui@surrey.ac.uk}
}

\begin{document}
\maketitle
\begin{abstract}
Generative AI (GenAI) is transforming creative workflows through the capability to synthesize and manipulate images via high-level prompts. Yet creatives are not well supported to receive recognition or reward for the use of their content in GenAI training. To this end, we propose ProMark, a causal attribution technique to attribute a synthetically generated image to its training data concepts like objects, motifs, templates, artists, or styles. The concept information is proactively embedded into the input training images using imperceptible watermarks, and the diffusion models (unconditional or conditional) are trained to retain the corresponding watermarks in generated images. We show that we can embed as many as $2^{16}$ unique watermarks into the training data, and each training image can contain more than one watermark. ProMark can maintain image quality whilst outperforming correlation-based attribution. Finally, several qualitative examples are presented, providing the confidence that the presence of the watermark conveys a causative relationship between training data and synthetic images.
\end{abstract}

\section{Introduction}
\label{sec:intro}

GenAI is able to create high-fidelity synthetic images spanning diverse concepts, largely due to advances in diffusion models, \eg~DDPM~\cite{ho2020denoising}, DDIM~\cite{meng2023distillation}, LDM~\cite{rombach2022high}.   
GenAI models, particularly diffusion models, have been shown to closely adopt and sometimes directly memorize the style and the content of different training images – defined as “concepts” in the training data~\cite{carlini2019secret, kumari2023ablating}. 
This leads to concerns from creatives whose work has been used to train GenAI. 
Concerns focus upon the lack of a means for attribution, \eg recognition or citation, of synthetic images to the training data used to create them and extend even to calls for a compensation mechanism (financial, reputational, or otherwise) for GenAI's derivative use of concepts in training images contributed by creatives.

\begin{figure}[t!]
\centering
\includegraphics[trim={0 -4 0 0},clip,width=1\columnwidth]{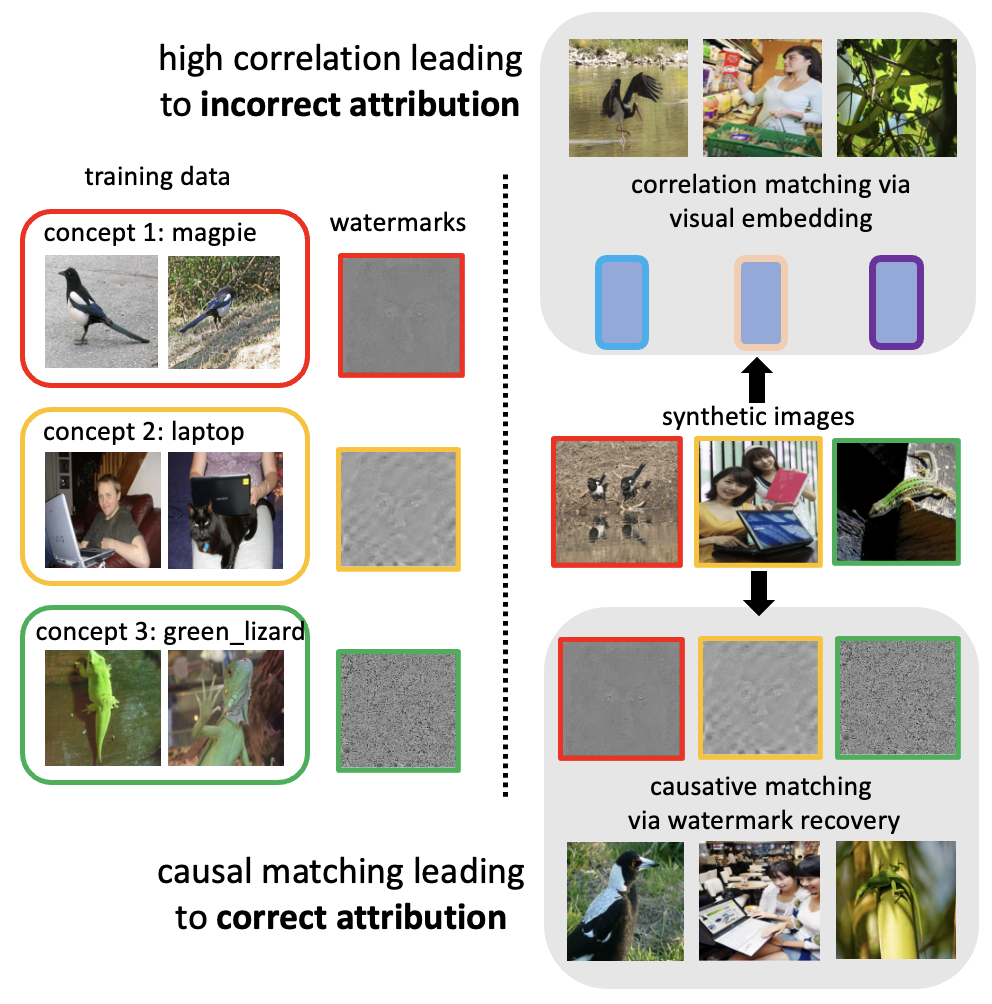}
\vspace{-4mm}
\caption{
\textbf{ Causative \vs correlation-based matching for concept attribution.}
\methodName identifies the training data most responsible for a synthetic image (`attribution').
Correlation-based matching doesn't always perform the data attribution properly. We propose \methodName, which is a proactive approach involving adding watermarks to training data and recovering them from the synthetic image to perform attribution in a causative way. 
\vspace{-4mm}}
\label{fig:teaser}
\end{figure}

We refer to this problem as {\em concept attribution} – the ability to attribute generated images to the training concept/s which have most directly influenced their creation.  
Several passive techniques have recently been proposed to solve the attribution problem~\cite{balan2023ekila, ruta2021aladin, wang2023evaluating}. 
These approaches use visual correlation between the generated image and the training images for attribution. 
Whilst they vary in their method and rationale for learning the similarity embedding – all use some forms of contrastive training to learn a metric space for visual correlation.

We argue that although correlation can provide visually intuitive results, a measure of similarity is not a causative answer to whether certain training data is responsible for the generation of an image or not.  Further, correlation-based techniques can identify close matches with images that were not even present in the training data.



Keeping this in mind, we explore an intriguing field of research which is developing around proactive watermarking methodologies~\cite{yeh2020disrupting,ruiz2020disrupting, wang2021faketagger,asnani2023malp}, 
that employ signals, termed \textit{templates} to encrypt input images before feeding them into the network. These works have integrated and subsequently retrieved templates to bolster the performance of the problem at hand. Inspired by these works, we introduce ProMark, a proactive watermarking-based approach for GenAI models to perform concept attribution in a causative way.  The technical contributions of ProMark are three-fold:
 

\noindent \textbf{1. Causal vs.~Correlation-based Attribution.}
ProMark performs causal attribution of synthetic images to the predefined concepts in the training images that influenced the generation. Unlike prior works that visually correlate synthetic images with training data, we make no assumption that visual similarity approximates causation. ProMark ties watermarks to training images and scans for the watermarks in the generated images, enabling us to demonstrate rather than approximate/imply causation.  This provides confidence in grounding downstream decisions such as legal attribution or payments to creators. 

\noindent \textbf{2. Multiple Orthogonal Attributions.} We propose to use orthogonal invisible watermarks to proactively embed attribution information into the input training data and add a BCE loss during the training of diffusion models to retain the corresponding watermarks in the generated images. We show that ProMark causatively attributes as many as $2^{16}$ unique training-data concepts like objects, scenes, templates, motifs, and style, where the generated images can simultaneously express one or two orthogonal concepts.

\noindent \textbf{3. Flexible Attributions.}
ProMark can be used for training conditional or unconditional diffusion models and even finetuning a pre-trained model for only a few iterations. We show that ProMark's causative approach achieves higher accuracy than correlation-based attribution over five diverse datasets (\cref{subsec:unconditional}): Adobe Stock, ImageNet, LSUN, Wikiart, and BAM while preserving synthetic image quality due to the imperceptibility of the watermarks.

\cref{fig:teaser} presents our scenario, where synthetic image(s) are attributed back to the most influential GenAI training images. Correlation-based techniques \cite{balan2023ekila,wang2023evaluating} try to match the high-level image structure or style. Here, the green-lizard synthetic image is matched to a generic green image without a lizard \cite{balan2023ekila}. With ProMark's causative approach, the presence of the green-lizard watermark in the synthetic image will correctly indicate the influence of the similarly watermarked concept group of lizard training images.

\section{Related Works}

\begin{table}[t]
\centering
\caption{\textbf{Comparison of \methodName with prior works}. Uniquely, we perform causative attribution using proactive watermarking to attribute multiple concepts.  [Keys: emb.: embedding, obj.: object, own.: ownership, sem.: semantic, sty.: style, wat.: watermark]}
\label{tab:rel_works}
\setlength\tabcolsep{0.05cm}
\begin{adjustbox}{width=1\columnwidth}
\begin{tabular}{c|c|c|c|c|c|c}
\hline\hline
\rowcolor{mygray} Method & Scheme & Task & Match & \# Class & Multiple & Attribution \\
\rowcolor{mygray} & type &  & type & & attribution & type \\\hline

\cite{ruta2021aladin} & passive & attribution & emb. & - & \xmark & sty.\\
\rowcolor{mygray}\cite{balan2023ekila} & passive & attribution & emb. & - & \xmark & obj.\\
\cite{wang2023evaluating} & passive & attribution & emb. & $693$ & \xmark & sty., obj.\\
\rowcolor{mygray}\cite{fernandez2023stable} & passive & detect & wat. & $2$ & \xmark & -\\
\cite{liu2023watermarking} & passive & detect & wat. & $2$ & \xmark & -\\
\rowcolor{mygray}\cite{cui2023diffusionshield} & passive & detect & wat. & $2$ & \xmark & -\\
\cite{wang2021faketagger} & proactive & detect & wat. & $2$ & - & -\\
\rowcolor{mygray}\cite{asnani2022proactive} & proactive & detect & wat. & $2$ & - & -\\
\cite{asnani2023malp} & proactive & localization & wat. & $2$ & - & -\\
\rowcolor{mygray}\cite{asnani2023probed} & proactive & obj. detect & - & $90$ & - & -\\\hline
& & & & & & sty., obj.\\
\multirow{-2}{*}{\methodName} & \multirow{-2}{*}{proactive} & \multirow{-2}{*}{attribution} & \multirow{-2}{*}{wat.} & \multirow{-2}{*}{$2^{16}$} & \multirow{-2}{*}{\cmark} & own., sem. \\
\hline\hline
\end{tabular}
\end{adjustbox}
\vspace{-4mm}
\end{table}
\minisection{Passive Concept Attribution.} 
Concept attribution differs from model~\cite{repmix} or camera~\cite{camera-trace-erasing} attribution  in that the task is to determine the responsible training data for a given generation.  Existing concept attribution techniques are passive – they do not actively modify the GenAI model or training data but instead, measure the visual similarity (under some definition) of synthetic images and training data to quantify attribution for each training image. 
EKILA~\cite{balan2023ekila} proposes patch-based perceptual hashing (visual fingerprinting \cite{nguyen2021,Black_2021_CVPR}) to match the style of the query patches to the training data for attribution.  Wang~\etal~\cite{wang2023evaluating} finetune semantic embeddings like CLIP, DINO, \etc for the attribution task.  Both~\cite{balan2023ekila} and \cite{wang2023evaluating} explore ALADIN~\cite{ruta2021aladin} for style attribution. 
 ALADIN is a feature representation for fine-grained style similarity learned using a weakly supervised approach. 

All these works are regarded as passive approaches as they take the image as an attribute by correlating between generated and training image styles. 
Instead, our approach is a proactive scheme that adds a watermark to training images and performs attribution in a causal manner (\cref{tab:rel_works}).

\minisection{Proactive Schemes.}
Proactive schemes involve adding a signal/perturbation onto the input images to benefit different tasks like deepfake tagging~\cite{wang2021faketagger}, deepfake detection~\cite{asnani2022proactive}, manipulation localization~\cite{asnani2023malp}, object detection~\cite{asnani2023probed}, \etc. 
Some works~\cite{yeh2020disrupting, ruiz2020disrupting} disrupt the output of the generative models by adding perturbations to the training data. Alexandre~\etal~\cite{sablayrolles2020radioactive} tackles the problem of training dataset attribution by using fixed signals for every data type. 
These prior works successfully demonstrate the use of watermarks to classify the content of the AI-generated images proactively. 
We extend the idea of proactive watermarking to perform the task of causal attribution of AI-generated images to influential training data concepts. 
Watermarking has not been used to trace attribution in GenAI before.

\begin{figure*}[t]
\centering
\includegraphics[trim={0 -4 0 0},clip,width=1\textwidth]{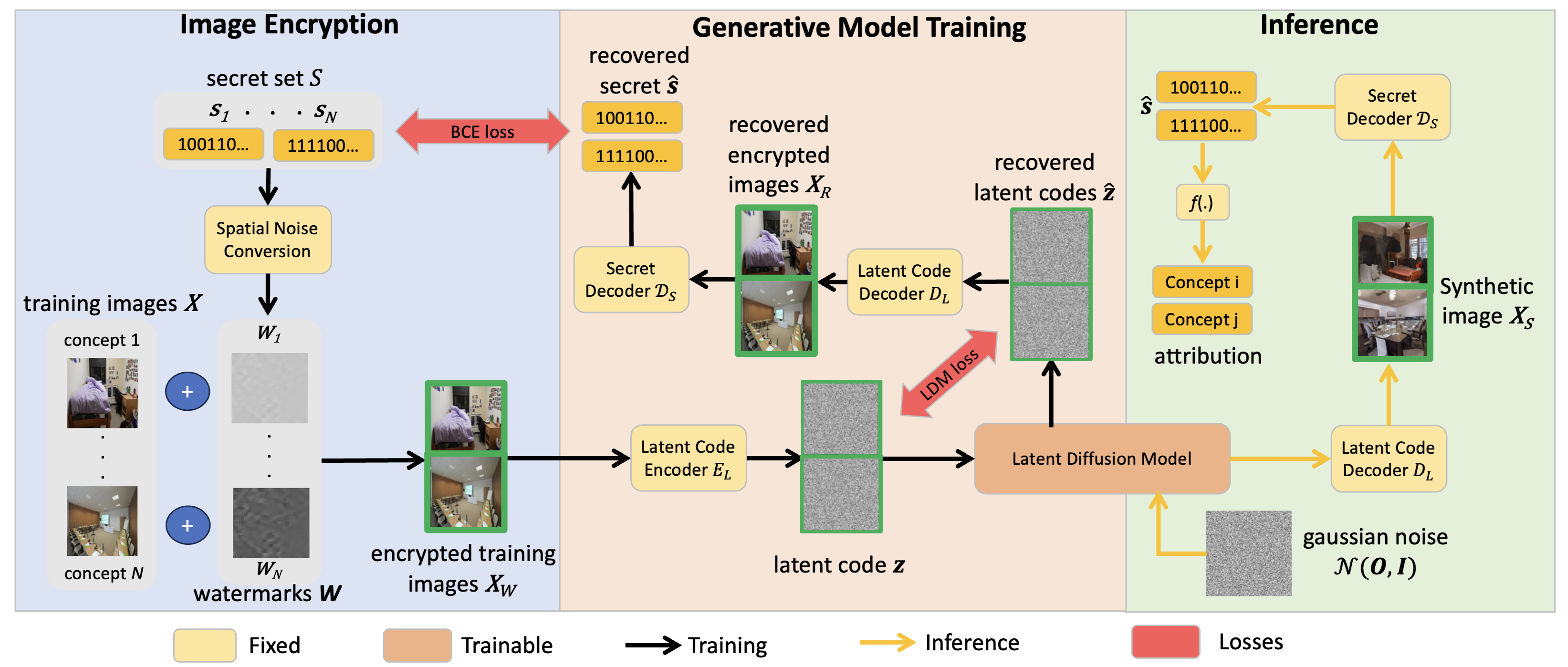}
\vspace{-5mm}
\caption{
\textbf{Overview of \methodName}. We show the training and inference procedure for our proposed method. Our training pipeline involves two stages, image encryption and generative model training. We convert the bit-sequences to spatial watermarks ($\vect{W}$), which are then added to the corresponding concept images ($\vect{X}$) to make them encrypted ($\vect{X}_W$). The generative model is then trained with the encrypted images using the LDM supervision. During training, we recover the added watermark using the secret decoder ($\mathcal{D}_S$) and apply the BCE supervision to perform attribution. To sample newly generated images, we use a Gaussian noise and recover the bit-sequences using the secret decoder to attribute them to different concepts. Best viewed in color.
\vspace{0mm}}
\label{fig:overview}
\vspace{-3mm}
\end{figure*}

\minisection{Watermarking of GenAI Models.}
It is an active research to watermark AI-generated images for the purpose of privacy protection. Fernandez1~\etal~\cite{fernandez2023stable}  fine-tune the LDM's decoder to condition on a bit sequence, embedding it in images for AI-generated image detection.
Kirchenbauer~\etal~\cite{kirchenbauer2023watermark} propose a watermarking method for language models by pre-selecting random tokens and subtly influencing their use during word generation. Zhao~\etal~\cite{zhao2023recipe} use a watermarking scheme for text-to-image diffusion models, while Liu~\etal~\cite{liu2023watermarking} verify watermarks by pre-defined prompts. \cite{cui2023diffusionshield, peng2023protecting} add a watermark for detecting copyright infringement. Asnani~\etal~\cite{asnani2023reverse} reverse engineer a fingerprint left by various GenAI models to further use it for recovering the network and training parameters of these models~\cite{reverse_eng_survey, asnani2023reverse}.
Finally, Cao~\etal~\cite{cao2023invisible} adds an invisible watermark for protecting diffusion models which are used to generate audio modality. 
Most of these works have used watermarking for protecting diffusion models, which enables them to add just one watermark onto the data. 
In contrast, we propose to add multiple watermarks to the training data and to a single image, which is a more challenging task than embedding a universal watermark. 

\section{Method}




\subsection{Background}
\minisection{Diffusion  Models.}
Diffusion models learn a data distribution $p(\vect{X}),\text{ where } \vect{X}\in \mathbb{R}^{h\times w \times 3}$ is the input image. 
They do this by iteratively reducing the noise in a variable that initially follows a normal distribution. 
This can be viewed as learning the reverse steps of a fixed Markov Chain with a length of $T$. 
Recently, LDM~\cite{rombach2022high} is proposed to convert images to their latent representation for faster training in a lower dimensional space than the pixel space. 
The image is converted to and from the latent space by a pretrained autoencoder consisting of an encoder $\vect{z}=\mathcal{E}_L(\vect{X})$ and a decoder $\vect{X}_R=\mathcal{D}_L(\vect{z})$, where $\vect{z}$ is the latent code and $\vect{X}_R$ is the reconstructed image. The trainable denoising module of the LDM is $\epsilon_{\theta}(\vect{z}_t, t); t = 1 . . . T$, where $\epsilon_{\theta}$ is trained to predict the denoised latent code $\hat{\vect{z}}$ from its noised version $\vect{z}_t$. This objective function can be defined as follows: 
\begin{equation}
    L_{LDM} = \mathbb{E}_{\mathcal{E}_L(\vect{X}), \epsilon \sim \mathcal{N}(0,1),t} \bigl[ ||\epsilon - \epsilon_{\theta}(\vect{z}_t,t)||^2_2\bigr],
    \label{eq:ldm_loss}
\end{equation}
where $\epsilon$ is the noise added at step $t$.

\minisection{Image Encryption.}  Proactive works~\cite{asnani2022proactive,asnani2023malp,asnani2023probed} have shown performance gain on various tasks by proactively transforming the input training images $\vect{X}$ with a watermark, resulting in an encrypted image. 
This watermark is either fixed or learned, depending on the task at hand. Similar to prior proactive works, our image encryption is of the form:
\begin{equation}
    \vect{X}_W = \mathcal{T}(\vect{X};\vect{W})={\vect{X}}+ m\times R(\vect{W},h,w),
    \label{eq:trans}
\end{equation}
where $\mathcal{T}$ is the transformation, $\vect{W}$ is the spatial watermark, $\vect{X}_W$ is the encrypted image, $m$ is the watermark strength, and $R(.)$ resizes $\vect{W}$ to the input resolution $(h,w)$.

We use the state-of-the-art watermarking technique RoSteALS~\cite{bui2023rosteals} to compute the spatial watermarks for encryption due to its robustness to image transformation and generalization (the watermark is independent of content of the input image). RoSteALS is designed to embed a secret of length $b$-bits into an image using robust and imperceptible watermarking. It comprises of a secret encoder $\mathcal{E}_S(\vect{s})$, which converts the bit-secret $\vect{s}\in \{0, 1\}^b$ into a latent code offset $\vect{z}_o$. It is then added to the latent code of an autoencoder $\vect{z}_w = \vect{z} + \vect{z}_o$. This modified latent code $\vect{z}_w$ is then used to reconstruct a watermarked image via autoencoder decoder. Finally, a secret decoder, denoted by $\mathcal{D}_S(X_W)$,  takes the watermarked images as input and predicts the bit-sequence $\hat{\vect{s}}$.

\subsection{Problem Definition}

Let $\mathcal{C} = \{c_1, c_2, \ldots, c_N\}$ be a set of $N$ distinct concepts within a dataset that is used for training a GenAI model for image synthesis. The problem of concept attribution can be formulated as follows:

{\it Given a synthetic image $\vect{X}_S$ generated by a GenAI model, the objective of concept attribution is to accurately associate $\vect{X}_S$ to a concept $c_i \in \mathcal{C}$ that significantly influenced the generation of $\vect{X}_S$.}

We aim to find a mapping $f: \vect{X}_S \rightarrow c_i$ such that
\begin{equation}
    c_i^* = \arg\max_{c_i \in \mathcal{C}} f(\vect{X}_S, c_i),
    \label{eq:att}
\end{equation}
where $c_i^*$ represents the concept most strongly attributed to image $\vect{X}_S$.


\subsection{Overview}
The pipeline of \methodName is shown in~\cref{fig:overview}. 
The principle is simple: if a specific watermark unique to a training concept can be detected from a generated image, it indicates that the generative model relies on that concept in the generation process. 
Thus, \methodName involves two steps: training data encryption via watermarks and generative model training with watermarked images. 

To watermark the training data, the dataset is first divided into $N$ groups, where each group corresponds to a unique concept that needs attribution. 
These concepts can be semantic (\eg objects, scenes, motifs or stock image templates) or abstract (\eg stylistic, ownership info). 
Each training image in a group is encoded with a unique watermark without significantly altering the image's perceptibility. 
Once the training images are watermarked, they are used to train the generative model. 
As the model trains, it learns to generate images based on the encrypted training images. 
Ideally, the generated images would have traces of watermarks corresponding to concepts they're derived from.

During inference, \methodName conforms to whether a generated image is derived from a particular training concept by identifying the unique watermark of that concept within the image. 
Through the careful use of unique watermarks, we can trace back and causally attribute generated images to their origin in the training dataset.

\subsection{Training}
During training, our algorithm is composed of two stages: image encryption and generative model training. 
We now describe each of these stages in detail.

\minisection{Image Encryption.}
The training data is first divided into $N$ concepts, and images in each partition are encrypted using a fixed spatial watermark $\vect{W}_j\in \mathbb{R}^{h\times w}$ ($j \in {0,1,2,...,N}$). Each noise $\vect{W}_j$ is a $b$-dim bit-sequence (secret) $\vect{s}_j = \{p_{j1}, p_{j2}, ..., p_{jb}\}$ where 
$p_{ji} \in \{0,1\}$. 

In order to compute the watermark $\vect{W}_j$ from the bit-sequence $\vect{s}_j$, we encrypt $100$ random images with $\vect{s}_j$ using pretrained RoSteALS secret encoder $\mathcal{E}_S(.)$ which takes $b=160$ length secret as input. 
From these encrypted images, we obtain $100$ noise residuals by subtracting the encrypted images from the originals, which are averaged to compute the watermark $\vect{W}_j$ as: 
\begin{equation}
    \vect{W}_j= \frac{1}{100} \sum_{i=1}^{100} (\vect{X}_i - \mathcal{E}_S(\vect{X}_i, \vect{s}_j)).
\end{equation}
The above process is defined as spatial noise conversion in~\cref{fig:overview}. The averaging of noise residuals across different images reduces the image content in the watermark and makes the watermark independent of any specific image. 
Additionally, the generated watermarks are orthogonal due to different bits for all $\vect{s}_j$, ensuring distinguishability from each other. 
With the generated watermarks, each training image is encrypted using~\cref{eq:trans} with one of the $N$ watermarks that correspond to the concept the image belongs to. 

\minisection{Generative Model Training.} 
Using the encrypted data, we train the LDM's denoising module $\epsilon_{\theta}(.)$ using the objective function (\cref{eq:ldm_loss}), where $\vect{z}_t$ is the noised version of:
\begin{equation}
    \vect{z} = \mathcal{E}_L(\vect{X}_{W_j}) = \mathcal{E}_L(\mathcal{T}(\vect{X};\vect{W}_j)),
\end{equation}
{\it i.e.}, the input latent codes $\vect{z}$ are generated using the encrypted images $\vect{X}_{W_j}$ for $j \in \{0,1,2....,N\}$. 

However, we found that only using LDM loss is insufficient to successfully learn the connection between the conceptual content and its associated watermark. 
This gap in learning presents a significant hurdle, as the primary aim is to trace back generated images to their respective training concepts via the watermark. 
To tackle this, an auxiliary supervision is introduced to LDM's training,
\begin{equation}
    L_{BCE}(\vect{s}_j, \hat{\vect{s}}) = -\frac{1}{b}\sum_{i=1}^{b} [ p_{ji} \log(\hat{p}_i) + (1-p_{ji}) \log(1-\hat{p}_i) ],
    \label{eq:binary_cross_entropy}
\end{equation}
where $L_{BCE}(.)$ is the binary cross-entropy (BCE) between the actual bit-sequence $\vect{s}_j$ associated with watermark $\vect{W}_j$ and the predicted bit-sequence $\hat{\vect{s}}$. The denoised latent code $\hat{\vect{z}}$ is then decoded using the autoencoder $\mathcal{D}_L(.)$, and the embedded secret $\hat{\vect{s}}$ is predicted by the secret decoder $\mathcal{D}_S(.)$ as:
\begin{equation}
    \hat{\vect{s}} = \mathcal{D}_S(\mathcal{D}_L(\hat{\vect{z}})).
    \label{eq:secret_extraction}
\end{equation}

By employing BCE, the model is guided to minimize the difference between the predicted watermark and the embedded watermark, hence improving the model's ability to recognize and associate watermarks with their respective concepts. Finally, our objective is to minimize the loss function $L_{attr} = L_{LDM} + \alpha L_{BCE}$ during training, where $\alpha$ is set to $2$ for our experiments.

\subsection{Inference}
\label{subsec:inference}
After the LDM learns to associate the watermarks with concepts, we use random Gaussian noise to sample the newly generated images from the model. While the diffusion model creates these new images, it also embeds a watermark within them. 
Each watermark maps to a distinctive orthogonal bit-sequence associated with a specific training concept, serving as a covert signature for attribution.

To attribute the generated images and ascertain which training concept influenced them, we predict the secret embedded by the LDM in the generated images (see~\cref{eq:secret_extraction}). Given a predicted binary bit-sequence, $\hat{\vect{s}} = \{\hat{p_1}, \hat{p_2}, ..., \hat{p_b}\}$ and all the input bit-sequences $\vect{s}_j$ for $j \in {0,1,2...,N}$, we define the attribution function, $f$, in~\cref{eq:att} as:
\begin{equation}
    f(\hat{\vect{s}}, \vect{s}_j) = \sum_{k=1}^{b} [\hat{p}_k = p_{jk}],
    \label{eq:similarity}
\end{equation}
where \( [\hat{p}_k = p_{ik}] \) acts as an indicator function, returning $1$ if the condition is true, {\it i.e.}, the bits are identical, and $0$ otherwise. 
Consequently, we assign the predicted bit sequence to the concept whose bit sequence it most closely mirrors — that is, the concept \( j^* \) for which $f(\hat{\vect{s}}, \vect{s}_{j^*})$ is maximized:
\begin{equation}
    j^* = \arg\max_{j \in \{1, 2, ..., N\}} f(\hat{\vect{s}}, \vect{s}_{j}).
    \label{eq:predicted_concept}
\end{equation}
\noindent In other words, the concept whose watermark is most closely aligned with the generated image's watermark is deemed to be the influencing source behind the generated image. 

\subsection{Multiple Watermarks}
\label{sec:multi-water}
In prior image attribution works, an image is usually attributed to a single concept (\eg image content or image style). However, in real-world scenarios, an image may encapsulate multiple concepts. This observation brings forth a pertinent question: ``Is it possible to use multiple watermarks for multi-concept attribution within a single image?"

In this paper, we propose a novel approach to perform multi-concept attribution by embedding multiple watermarks into a single image. 
In our preliminary experiments, we restrict our focus to the addition of two watermarks. 
To achieve this, we divide the image into two halves and resize each watermark to fit the respective halves. 
This ensures that each half of the image carries distinct watermark information pertaining to a specific concept.

For the input RGB image $\vect{X}$, \{$\vect{W}_i,\vect{W}_j$\} are the watermarks for two secrets \{$\vect{s}_i, \vect{s}_j$\}, we formulate the new transformation $\mathcal{T}$ as:
\begin{align*}
    \mathcal{T}(\vect{X}; \vect{W}_i, \vect{W}_j)= &\Big\{ \vect{X}_{left}, \vect{X}_{right}\Big \}\\
    =&\Big \{(\vect{X}(:,0:\frac{w}{2},:)+R(\vect{W}_i, h, \frac{w}{2})), \\
    &(\vect{X}(:, \frac{w}{2}:w,:)+R(\vect{W}_j, h,\frac{w}{2}) \Big \},
\end{align*}
where $\{.\}$ is the horizontal concatenation. The loss function uses the two predicted secrets ($\hat{\vect{s}}_1$ and $\hat{\vect{s}}_2$) from the two halves of the generated image, defined as:
\begin{align*}
    L_{attr} = L_{LDM} + \alpha(L_{BCE}(\vect{s}_i, \hat{\vect{s}}_1) + L_{BCE}(\vect{s}_j, \hat{\vect{s}}_2)).
\end{align*}

\section{Experiments}

\subsection{Unconditional Diffusion Model}
\label{subsec:unconditional}
In this section, we train multiple versions of unconditional diffusion models~\cite{rombach2022high} to demonstrate that \methodName can be used to attribute a variety of concepts in the training data. In each case, the model is trained starting from random initialization of LDM weights. Described next are details of the datasets and evaluation protocols. 

\minisection{Datasets}
We use $5$ datasets spanning attribution categories like image templates, scenes, objects, styles, and artists. For each dataset, we consider the dataset classes as our attribution categories. For each class in a dataset, we use $90\%$ images for training, and $10\%$ for evaluation, unless specified otherwise.
\begin{enumerate}
    \item Stock: We collect images from Adobe Stock, comprising of near-duplicate image clusters like templates, symbols, icons,~\etc. An example image from some clusters is shown in the supplementary. We use $100$ such clusters, each with $2K$ images. 

    \item LSUN: The LSUN dataset~\cite{yu2015lsun} comprises $10$ scene categories, such as bedrooms and kitchens. 
    It's commonly used for scene classification, training generative models like GANs, and anomaly detection. Same as the Stock dataset, we use $2K$ images per class. 

    \item Wiki-S: The WikiArt dataset~\cite{artgan2018} is a collection of fine art images spanning various styles and artists. We use the $28$ style classes with $580$ average images per class. 

    \item Wiki-A: From the WikiArt dataset~\cite{artgan2018} we also use the $23$ artist classes with $2,112$ average images per class.

    \item ImageNet: We use the ImageNet dataset~\cite{deng2009imagenet} which comprises of $1$ million images across $1K$ classes. For this dataset, we use the standard validation set with $50K$ for evaluation and the remaining images for training. 
\end{enumerate}

\begin{table}[t]
\rowcolors{1}{mygray}{white}
\centering
\small
\caption{Comparison with prior works for unconditional diffusion model on various datasets. [Keys: str.: strength]}
\begin{adjustbox}{width=1.0\columnwidth}
\label{tab:uncond_sota}
\begin{NiceTabular}{c|c|c|c|c|c|c}
\hline\hline
\rowcolor{mygray} & Str. & \multicolumn{5}{c}{Attribution Accuracy (\%) $\bf \uparrow$} \\\hhline{~|~|-|-|-|-|-}
\rowcolor{mygray} \multirow{-2}{*}{Method} & (\%) & Stock & LSUN & Wiki-A & Wiki-S & ImageNet \\\hline
ALADIN~\cite{ruta2021aladin} & - & $99.86$ & $46.27$ & $48.95$ & $33.25$ & $9.25$\\
\rowcolor{mygray}CLIP~\cite{radford2021learning} & - & $75.67$& $87.13$ & $77.58$ & $60.84$ & $60.12$\\
F-CLIP~\cite{wang2023evaluating} & - & $78.49$ & $87.39$ & $77.23$ & $60.43$ & $62.83$ \\
\rowcolor{mygray} SSCD~\cite{pizzi2022self} & - & $99.63$ & $73.26$ & $69.51$ & $50.37$ & $37.32$  \\
EKILA~\cite{balan2023ekila} & - & $99.37$& $70.60$ & $51.23$ & $37.06$ & $38.00$\\\hline
\rowcolor{mygray} & $30$ & $100$ & $95.12$ & $97.45$ & $98.12$ & $83.06$ \\\hhline{~|-|-|-|-|-|-|-}
\rowcolor{mygray}\multirow{-2}{*}{\methodName} & $100$ & $\bf 100$ &  $ \bf 100$ &  $\bf 100$ & $\bf 100$ &  $\bf 91.07$\\\hline\hline
\end{NiceTabular}
\end{adjustbox}
\vspace{-4mm}
\end{table}

\begin{figure}[t!]
\centering
\includegraphics[trim={0 -4 0 0},clip,width=1\columnwidth]{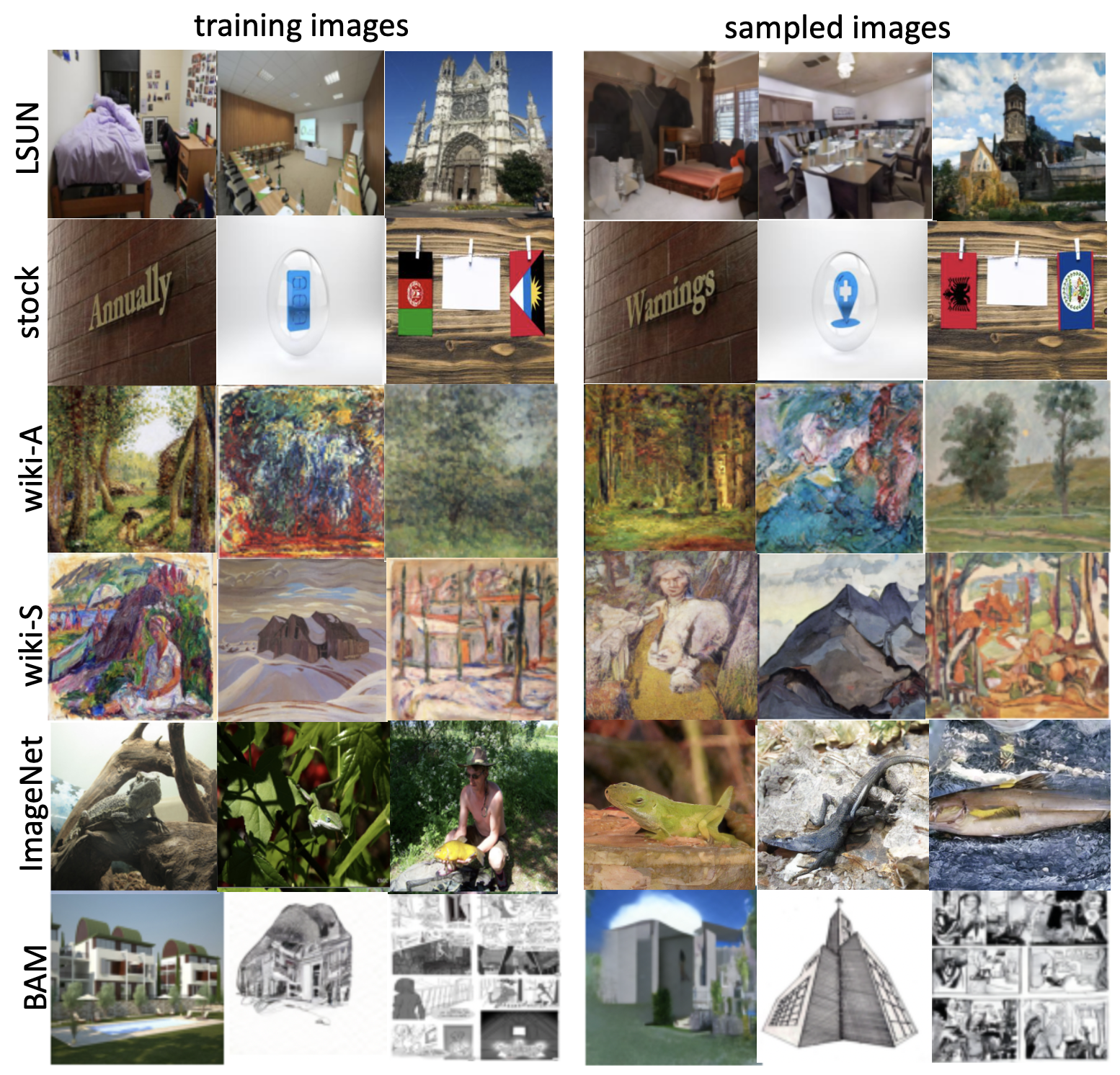}
\vspace{-6mm}
\caption{
\textbf{Example training and newly sampled images of different datasets for the corresponding classes}. We observe a similar content in the inference image compared with the training image of the predicted class. 
\vspace{-4mm}}
\label{fig:uncon_samples}
\end{figure}

\minisection{Evaluation Protocol} 
For all datasets, the concept attribution performance is tested on the held-out data as follows. For a held-out image, we first encrypt it with the concept's watermark. Then using the latent code of the encrypted image, we noise it till a randomly assigned timestamp and apply our trained diffusion model to reverse back to the initial timestamp with the estimated noise. The denoised latent code is then decoded using the autoencoder $\mathcal{D}_L(.)$, and the embedded secret is predicted using the secret decoder $\mathcal{D}_S(.)$. Using~\cref{eq:predicted_concept}, we compute the predicted concept and calculate the accuracy using the ground-truth concept.

\minisection{Results} 
Shown in~\cref{tab:uncond_sota} is the attribution accuracy of \methodName at two watermark strengths \ie~$100\%$ and $30\%$ which is set by variable $m$ in~\cref{eq:trans}. \methodName outperforms prior works, achieving near-perfect accuracy on all the datasets when the watermark strength is $100\%$. However, the watermark introduces visual artifacts~\cite{bui2023rosteals} if the watermark strength is full. Therefore, we decrease the watermark strength to $30\%$ before adding it to the training data (see~\cref{subsec:ablation} for ablation on watermark strength). Even though our performance drops at a lower watermark strength, we still outperform the prior works. This shows that our causal approach can be used to attribute a variety of concepts in the training data with an accuracy higher than the prior passive approaches. 

\cref{fig:uncon_samples} (rows 1-5) shows the qualitative examples of the newly sampled images from each of the trained models. For each model, we sample the images using random Gaussian noise until we have images for every concept. The concept for each image is predicted using the secret embedded in the generated images. Shown in each row of~\cref{fig:uncon_samples} are three training images (columns 1-3) and three sampled images from the corresponding concepts (columns 4-6). This shows that \methodName makes the diffusion model embed the corresponding watermark for the class of the generated image, thereby demonstrating the usefulness of our approach.

\begin{figure}[t!]
\centering
\includegraphics[trim={0 -4 0 0},clip,width=1\columnwidth]{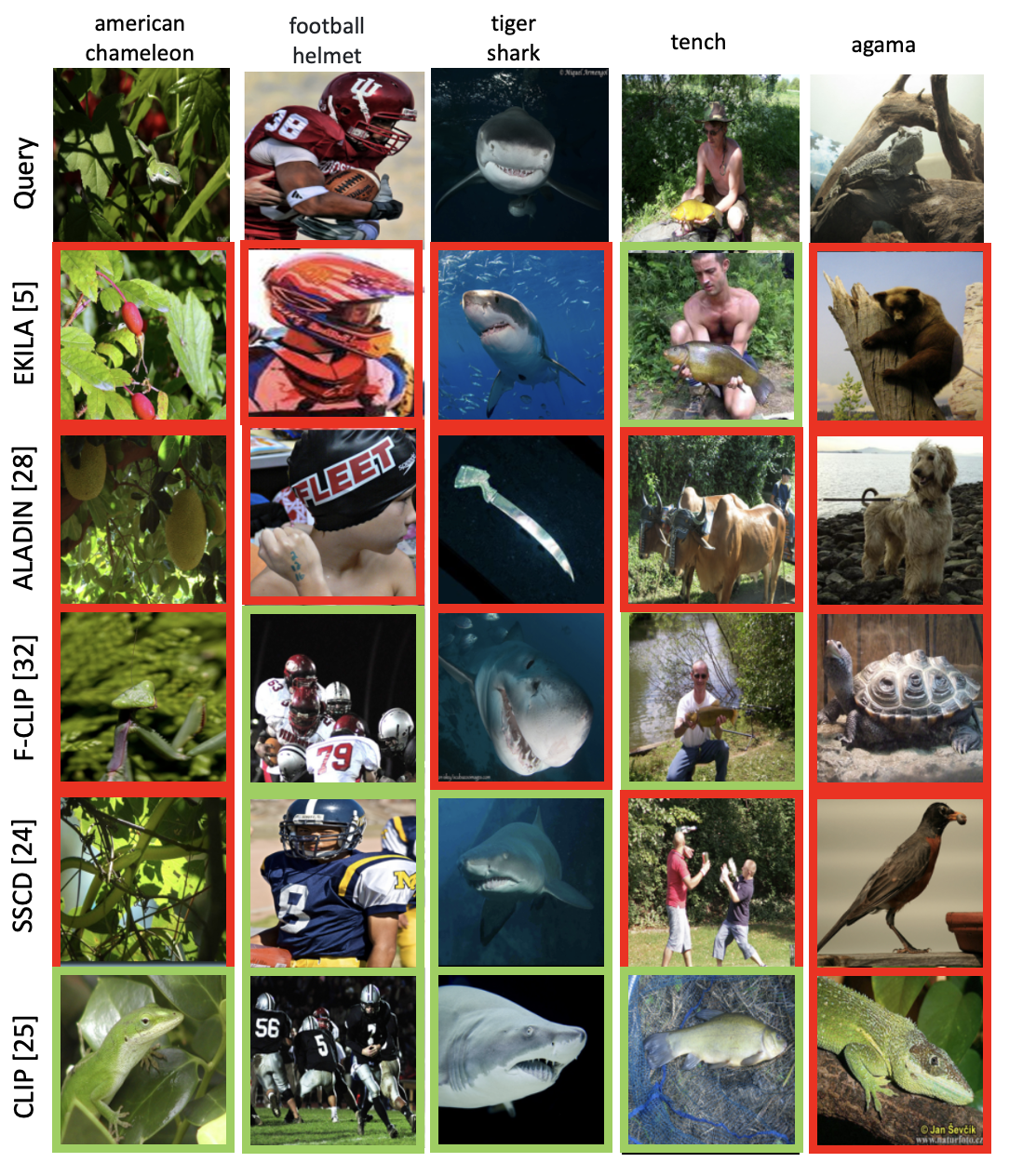}
\vspace{-6mm}
\caption{
\textbf{Visual results of prior embedding-based works.} We show the image of the closest matched embedding for each method on ImageNet. We highlight images green for correct attribution, otherwise red.   Embedding-based works do not always attribute to the correct concept. 
\vspace{-4mm}}
\label{fig:visual_base_comp}
\end{figure}

Shown in~\cref{fig:visual_base_comp} are the nearest images retrieved using the embedding-based methods (row ($2$)-($6$)) for the query images from the ImageNet (row ($1$)). For each image retrieval, we highlight the correct/incorrect attribution using a green/red box. As we can see, the correlation-based prior techniques rely on visual similarity between the query and the retrieved images, ignoring the concept. However, for each query image, \methodName predicts the correct concept corresponding to the query image (\cref{fig:uncon_samples}). 

\subsection{Multiple Watermarks}

We evaluate the effectiveness of \methodName for multi-concept attribution. As before, an unconditional diffusion model is trained starting from random initialization, and each image in the training data is encrypted with two watermarks as outlined in \cref{sec:multi-water}. 

\minisection{Dataset} For this experiment, we use the BAM dataset~\cite{wilber2017bam}, comprising contemporary artwork sourced from Behance, a platform hosting millions of portfolios by professionals and artists. This dataset uniquely categorizes each image into two label types: media and content. It encompasses $7$ distinct labels for media and $9$ for content, culminating in a diverse set of $63$ label pairs, with $4,593$ average images in these label pairs. For each class pair, we use $90\%$ data for training and $10\%$ for held-out evaluation. 

\begin{table}[t]
\rowcolors{1}{mygray}{white}
\centering
\small
\caption{Multi-concept attribution comparison with baselines.}
\begin{adjustbox}{width=1.0\columnwidth}
\label{tab:cond_mul_sota}
\begin{NiceTabular}{c|c|c|c|c}
\hline\hline
\rowcolor{mygray} & Strength & \multicolumn{3}{c}{Attribution Accuracy (\%) $\bf \uparrow$} \\\hhline{~|~|-|-|-}
\rowcolor{mygray} \multirow{-2}{*}{Method} &  (\%) & Media & Content & Combined\\\hline
ALADIN~\cite{ruta2021aladin} & - & $42.16$ & $41.25$ & $34.97$\\
\rowcolor{mygray}CLIP~\cite{radford2021learning} & - & $46.71$ & $45.12$ & $42.36$\\
F-CLIP~\cite{wang2023evaluating} & - & $52.12$ & $51.56$ & $46.23$\\
\rowcolor{mygray} SSCD~\cite{pizzi2022self} & - & $47.06$ & $46.09$ & $40.61$\\
EKILA~\cite{balan2023ekila} & - & $43.72$ & $43.58$ & $37.09$\\\hline
\rowcolor{mygray} \methodName (single) & 30 & - & - & $\bf 97.73$\\
 & $30$ & $91.33$ & $89.21$ & $84.66$\\\hhline{~|-|-|-|-}
\multirow{-2}{*}{\methodName (multi)} & $ 100$ & $\bf95.61$ & $\bf93.31$ & $90.12$\\\hline\hline
\end{NiceTabular}
\end{adjustbox}
\vspace{-3mm}
\end{table}

\minisection{Results} The same evaluation is performed as described in~\cref{subsec:unconditional}, except the accuracy is now computed for two concepts instead of one. Shown in~\cref{tab:cond_mul_sota} is the attribution accuracy for the two concepts individually and simultaneously. To benchmark the effectiveness of \methodName, we also compare against baselines, where \methodName outperforms baselines, achieving a combined attribution accuracy of $90.12\%$ as compared to $46.61\%$ for F-CLIP~\cite{wang2023evaluating}. 
We believe our findings substantiate that \methodName can be extended to a scenario where the generated images are composed of several unique concepts from the training images. For ablation,  we train \methodName with $7\times8$ classes, with each pair of media and content as an individual concept. \methodName is able to achieve $97.73\%$ attribution accuracy for single-concept, higher than the performance achieved for multi-concept case \ie $90.12\%$. However, single concept approach is not scalable when the number of concepts in an image increases, as the number of watermarks would grow exponentially ($7\times8$ \vs $7+8$). Therefore, transitioning to a multi-concept scenario is more appropriate for real-world scenarios, where scalability and practicality are crucial.

In the final row of~\cref{fig:uncon_samples}, we present qualitative examples of newly sampled images from the model trained on the BAM dataset. Observations indicate that these sampled images successfully adopt both media and content corresponding to training images of the same concept. This provides empirical evidence of \methodName's effectiveness in facilitating multi-concept attribution.

\subsection{Number of Concepts}
\begin{figure*}[t!]
\centering
\includegraphics[trim={0 -4 0 0},clip,width=1\textwidth]{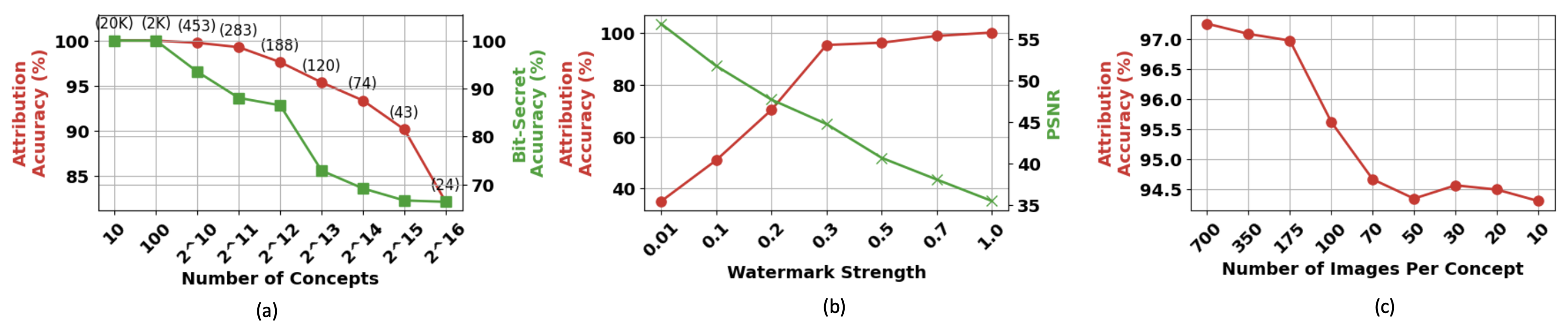}
\vspace{-8mm}
\caption{\textbf{Ablation experiments}: We show the results for ablating multiple parameters of \methodName. (a) Number of concepts, (b) watermark strength, and (c) number of images per concept. }
\vspace{-4mm}
\label{fig:ablation}
\end{figure*}

AI models leverage large-scale image datasets~\cite{rombach2022high, radm, ho2020denoising, meng2023distillation}, encompassing a broad spectrum of concepts. This diversity necessitates concept attribution methods that can maintain high performance across numerous concepts. 
In this context, we test \methodName with an exponentially increasing number of concepts. 
Our dataset comprises Adobe Stock images with near duplicate image templates (used as concepts).
As we escalate the number of concepts, we concurrently reduce the per-concept image count, only $24$ images per concept for $2^{16}$ concepts, see the red curve of~\cref{fig:ablation}~(a) for image count. This is done to obtain balanced image distribution and also to challenge \methodName's robustness.

\begin{table}[t]
\rowcolors{1}{mygray}{white}
\centering
\small
\caption{Comparison with different baselines for the conditional model trained on ImageNet dataset.}
\begin{adjustbox}{width=0.9\columnwidth}
\label{tab:conditional-imagenet}
\begin{NiceTabular}{c|c|c|c}
\hline\hline
\rowcolor{mygray} & Strength & \multicolumn{2}{c}{Attribution Accuracy (\%)  $\bf \uparrow$} \\\hhline{~|~|-|-}
\rowcolor{mygray} \multirow{-2}{*}{Method} & (\%) & Held-out data & New images\\\hline
ALADIN~\cite{ruta2021aladin} & - & $9.25$ & $0.18$\\
\rowcolor{mygray}CLIP~\cite{radford2021learning} & - & $60.12$ & $41.01$\\
F-CLIP~\cite{wang2023evaluating} & - & $62.83$ & $50.19$\\
\rowcolor{mygray} SSCD~\cite{pizzi2022self} & - & $37.32$ & $30.10$\\
EKILA~\cite{balan2023ekila} & - & $38.00$ & $29.06$\\\hline
\rowcolor{mygray} & $30$ & $91.24$ & $87.30$\\\hhline{~|-|-|-}
\rowcolor{mygray}\multirow{-2}{*}{\methodName} & $ 100$ & $\bf 95.60$ & $\bf 90.13$\\\hline\hline
\end{NiceTabular}
\end{adjustbox}
\vspace{-4mm}
\end{table}

The outcomes, depicted in~\cref{fig:ablation}(a) red curve, indicate an anticipated decline in \methodName's efficacy in line with the increase in the number of concepts, reducing from $100\%$ attribution accuracy for $10$ concepts (chance accuracy $10\%$) to $82\%$ for $2^{16}$ concepts (chance accuracy $1.5e$-$3\%$). This reduction in attribution accuracy is correlated with the reduction in bit-secret accuracy (green curve) for every predicted secret, indicating poor watermark recovery due to the increased confusion between the watermarks. Notwithstanding the increased difficulty, \methodName demonstrates commendable performance, underscoring its potential in real-world applications. 

\subsection{Conditional Diffusion Model}
As the diffusion models are usually trained with conditions to guide generation, we also evaluate using the conditional LDM model~\cite{rombach2022high}. For this, we fine-tune a model pretrained of the ImageNet dataset (see~\cref{subsec:unconditional}), where the $1000$ ImageNet classes are used as model conditions and also as the $1000$ concepts.

\minisection{Evaluation Protocol}
In addition to the evaluation on the held-out data (see~\cref{subsec:unconditional}), we also perform the quantitative evaluation on the newly sampled images as follows. We use the labels of the ImageNet dataset as conditions to sample $10K$ images ($10$ images per label). Using these labels as the ground-truth concept for a newly sampled image, we compute the accuracy of the concept predicted by the embedded watermark in the generated images. 

\minisection{Results}
The accuracies for held-out and newly sampled images are shown in~\cref{tab:conditional-imagenet}. The performance on the held-out dataset for the conditional model improves compared to the unconditional models as the label conditions provide improved supervision for correct watermarks. \methodName also outperforms prior embedding-based works by a large margin on both held-out and newly sampled images. The attribution accuracy on the new images, however, is less than the held-out data. We hypothesize that it is because newly sampled images may contain more than one concept and can be more confusing to attribute. The high accuracy, even for newly sampled images, suggests that \methodName exhibits higher generalizability to unseen synthetic images.

\subsection{Ablation Study}
\label{subsec:ablation}
For the ablation experiment, we use Stock dataset with a varying number of concepts, and we train unconditional LDM models from random initialization.

\minisection{Strength of Watermark.}
The hyperparameter $m$ in~\cref{eq:trans} modulates the intensity of the watermark applied to the training images, ensuring encrypted images retain high quality. We systematically alter $m$ to examine its impact on the LDM's performance and the Peak Signal-to-Noise Ratio (PSNR) of the output images with reference to the held-out encrypted images. 
\cref{fig:ablation}(b) shows that attribution accuracy improves with increased $m$, plateauing beyond a threshold of $0.5$. The discernible compromise in image quality, as evidenced by the inverse relationship between intensity and PSNR, can be attributed to the use of fixed watermarks obtained using RoSteALS~\cite{bui2023rosteals}, which is originally optimized for robustness. In light of this, we select an optimal watermark strength of $0.3$, which balances between performance and PSNR. We measured the FID between original and newly sampled images from a pretrained ImageNet conditional model (trained without watermark) and ProMark model (trained with watermark), which is $13.28$ and $17.63$ respectively. This small increment shows negligible quality loss in the generated images due to \methodName.


\minisection{Number of Images Per Concept.}
To ascertain the optimal number of images required per concept for effective watermark learning, we ablate by fixing the number of concepts to $500$ and varying the number of images used to train the LDM. 
\cref{fig:ablation}(c) reveals that performance drops by $2.5\%$ when image count per concept is reduced from $700$ to $10$. 
Remarkably, the general efficacy of \methodName remains consistently high, suggesting a low sensitivity to the image count per concept. 
These results demonstrate that \methodName can successfully learn watermarks with as few as $10$ images per concept, highlighting its efficiency and potential for applications with limited data availability.

\minisection{Framework Design.}
\methodName employs BCE loss to instruct the LDM model in the accurate embedding of bit-sequence watermarks within generated images. 
The attribution performance degrades to $2\%$ when BCE loss is not used as compared to $100\%$ in~\cref{tab:uncond_sota}. This shows that removing BCE loss significantly impairs the LDM's performance, underscoring the necessity of this supervision in helping LDM embed watermarks effectively.

Also, \methodName incorporates a secret decoder to retrieve secret bit-sequence from synthesized images, rendering the process contingent upon the pretrained secret decoder. 
In contrast, prior works~\cite{asnani2022proactive, asnani2023malp, asnani2023probed} recover watermarks by training a dedicated decoder with the main model in an end-to-end fashion. 
To ablate this alternative approach, we train a standard decoder along with LDM by optimizing for the cosine similarity between the embedded and extracted watermarks. 
We see a degradation in performance from $100\%$ to $80.56\%$, indicating that the pretrained secret decoder is a better choice for our approach. This is due to the increased complexity of predicting watermarks of resolution $256^2$ as compared to $160$-bit sequence from the encrypted images.
\section{Conclusion}
We introduce a novel proactive watermarking-based approach, \methodName, for causal attribution. We use predefined training concepts like styles, scenes, objects, motifs, \etc. to attribute the influence of training data on generated images. We show \methodName's is effective across various datasets and model types, maintaining image quality while providing more accurate attribution on a large number of concepts. Our approach can also be extended to multi-concept attribution by embedding multiple watermarks onto the image. Finally, for each experiment, our approach achieves a higher attribution accuracy than the prior passive approaches. Such  attribution offers opportunities to recognize and reward creative contributions to generative AI, underpinning new models for value creation in the future creative economy \cite{CGA}.

\minisection{Limitations.}
In evaluating \methodName, we note a trade-off between image quality and attribution accuracy, which may need us to learn watermarks for attribution task. Our model is currently trained with predefined concepts and further research is needed on training paradigm when new concepts are introduced. While we use orthogonal watermarks for varied concepts like motifs and styles, this may not accurately reflect the interrelated nature of some concepts, suggesting another opportunity for future research. Finally, our results are specific to the LDM, and extending this approach to other GenAI models could provide a better understanding of \methodName's effectiveness.


{
    \small
    \bibliographystyle{ieeenat_fullname}
    \bibliography{main}
}

\twocolumn[\centering \section*{\Large \textbf{ProMark: Proactive Diffusion Watermarking for Causal Attribution \\ -- Supplementary material --\\[1cm]}}] 

\section{Multiple Watermark Configurations}

We investigate the application of dual watermarks, each positioned on opposing sides of the image. 
This exploration raises a pivotal query: ``Is the spatial positioning of watermarks critical to the performance?" 
To answer this, we ablate four distinct watermark configurations. 
As shown in~\cref{tab:multi-water-config}, there is a consistent performance across all watermark placements (left, right, top, bottom), thereby substantiating the spatial robustness of \methodName in watermark positioning.
\begin{table}[t]
\rowcolors{1}{mygray}{white}
\centering
\small
\caption{Multi-concept attribution performance across different configurations.}
\begin{adjustbox}{width=1.0\columnwidth}
\label{tab:multi-water-config}
\begin{NiceTabular}{c|c|c|c|c}
\hline\hline
\rowcolor{mygray} \multicolumn{2}{c}{Configuration} & \multicolumn{3}{c}{Attribution Accuracy (\%) $\bf \uparrow$} \\\hhline{-|-|-|-|-}
\rowcolor{mygray} Secret $1$ & Secret $2$ & Secret $1$ & Secret $2$ & Combined\\\hline
Left & Right & $95.61$ & $93.31$ & $90.12$\\
\rowcolor{mygray}Right & Left & $95.52$ & $93.35$ & $90.19$\\
Top & Bottom & $95.66$ & $93.70$ & $90.01$ \\
\rowcolor{mygray}Bottom & Top & $95.02$ & $93.46$ & $90.73$ \\\hline\hline
\end{NiceTabular}
\end{adjustbox}
\vspace{-2mm}
\end{table}

\section{Watermark Robustness}
We test our method against $14$ different degradations (blur, various noises, fog, etc.), by adopting the evaluation protocol detailed in the RoSteALS~\cite{bui2023rosteals}. We use $50$ watermarked training images from LSUN dataset and use unconditional LDM with a strength of $30\%$. The average attribution accuracy for training and generated images across all $14$ attacks is $90.21\pm 7.63\%$ and $89.51 \pm 8.18\%$, as compared to $95.12\%$ without any degradation, showing the robustness of our approach to multiple forms of watermark attack.

\section{Possibility of Concept Leakage}
We present multiple results where we attribute the images generated using non-watermarked data, for example via random latent code and conditional generation. We detect no retention of the watermark after noising or in random latent codes, with watermark detection accuracy of $50.56\%$ (chance $50\%$) after noising for $\geq900$ timestamps or in random latent codes. The LDM generates an image from noise through inversion, and the watermark is added during this GenAI model inference process. Our decoder is employed independently to identify the concept. To prove this, we evaluate our model in Table $4$ (main paper) for two more baselines, using held-out images ($1$) with no watermark encryption, and ($2$) encrypted with a different concept’s watermark. ProMark is able to attain an attribution accuracy of $94.32\%$ and $94.01\%$ respectively when evaluated with ground-truth concept watermark for both baselines compared to $95.60\%$ reported for watermarked held-out data. Therefore, when inverting generating images that encrypt no watermark, or encrypt incorrect watermark, the correct concept watermark is encrypted. 

\section{Computational Efficiency}

We demonstrate the computation efficiency of ProMark during inference (running watermark decoder to perform causal attribution), which costs $5.6$ms on one A$100$ GPU.  Training with watermarked data adds negligible cost to generative model training.  This is comparable to running inference on CLIP, or ALADIN to perform correlation based attribution ($28.32$ ms) but the additional cost of the embedding search is $87.91$ ms for a dataset of $20$K LSUN training images. ProMark therefore offers the advantage of both efficiency and causality for training data attribution. We will add this to the paper.

\section{Additional Watermark Strength Analysis}
\begin{figure}[t!]
\centering
\includegraphics[trim={0 -4 0 0},clip,width=0.9\columnwidth]{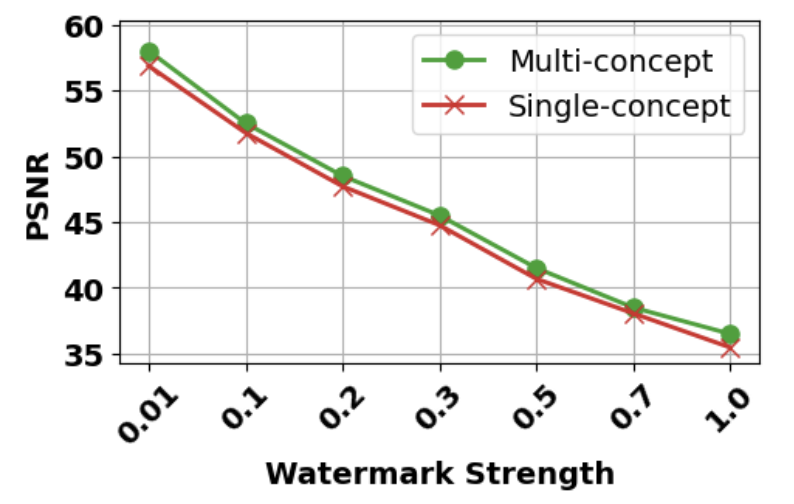}
\vspace{-6mm}
\caption{PSNR~\vs watermark strength for single vs multi-concept attribution. }
\label{fig:psnr_multi}
\vspace{-2mm}
\end{figure}

Our research introduces a new paradigm in concept attribution for images classified under multiple concepts.  We show the analysis of PSNR variation with watermark strength for the case of multi-concept attribution. The results are shown in~\cref{fig:psnr_multi}. Our findings indicate that, compared to single watermark cases, the PSNR for multi-concept images is marginally higher at equivalent watermark strengths. However, as expected, an increase in watermark strength generally leads to a decrease in PSNR.

\begin{figure*}[t!]
\centering
\includegraphics[trim={0 -4 0 0},clip,width=1\textwidth]{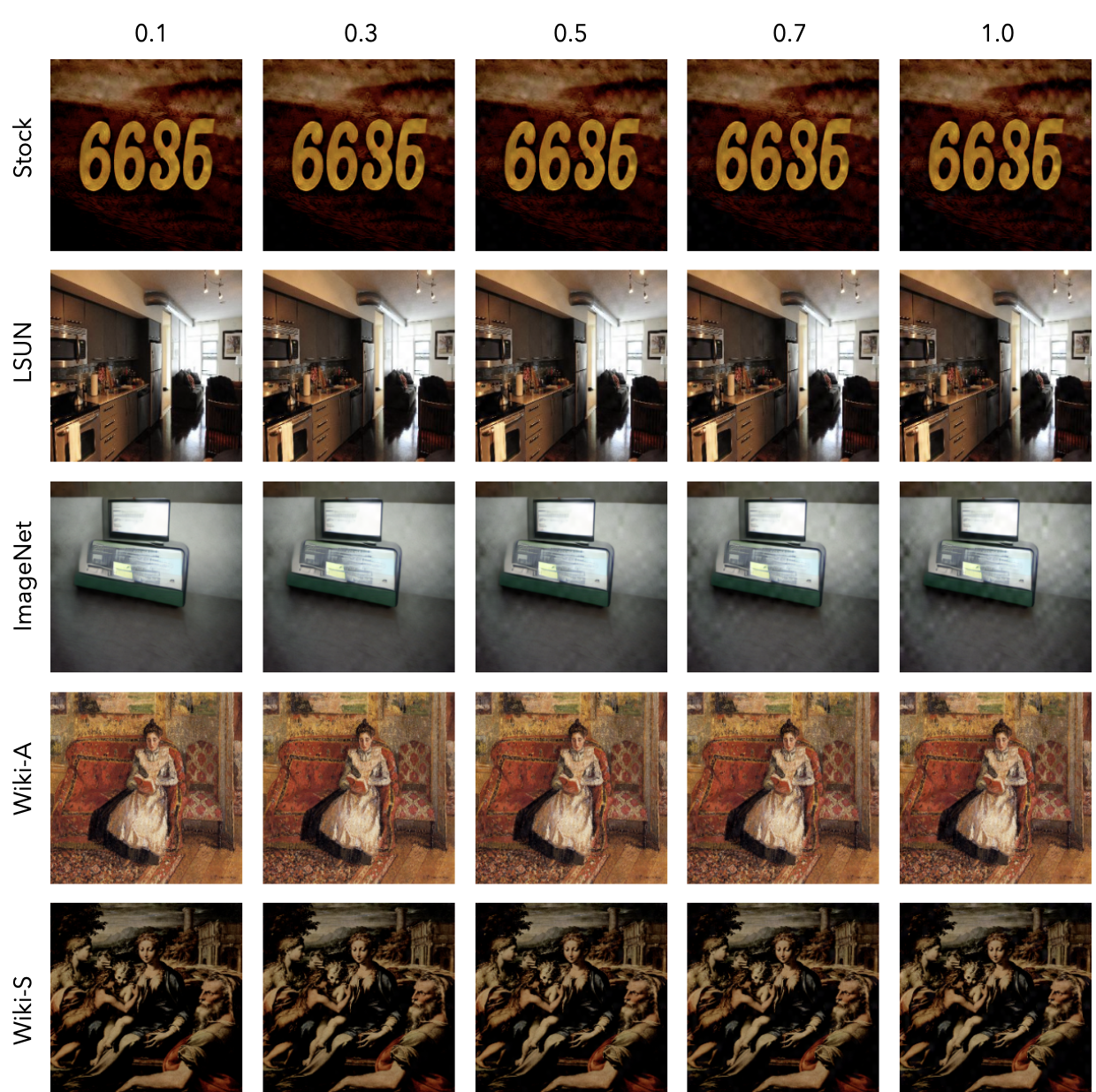}
\caption{Noise Strength visualization for different watermark strength }
\label{fig:noise_strength}
\end{figure*}

Furthermore, we have visualized images from different datasets to showcase the extent of degradation caused by varying watermark strengths. As discussed in Sec.~$4.5$, the performance of our method improves with increased watermark strength. Nevertheless, this increase in strength leads to a decline in image quality, evidenced by the emergence of bubble-like artifacts in the images, as shown in~\cref{fig:noise_strength} (the watermark strength ranges from $0.1$ to $1.0$).

\section{Watermark Discussion}
\begin{figure*}[t!]
\centering
\begin{tabular}{ccc}
Spatial Domain & Fourier Domain & Correlation Matrix \\
     \includegraphics[width=0.3\textwidth]{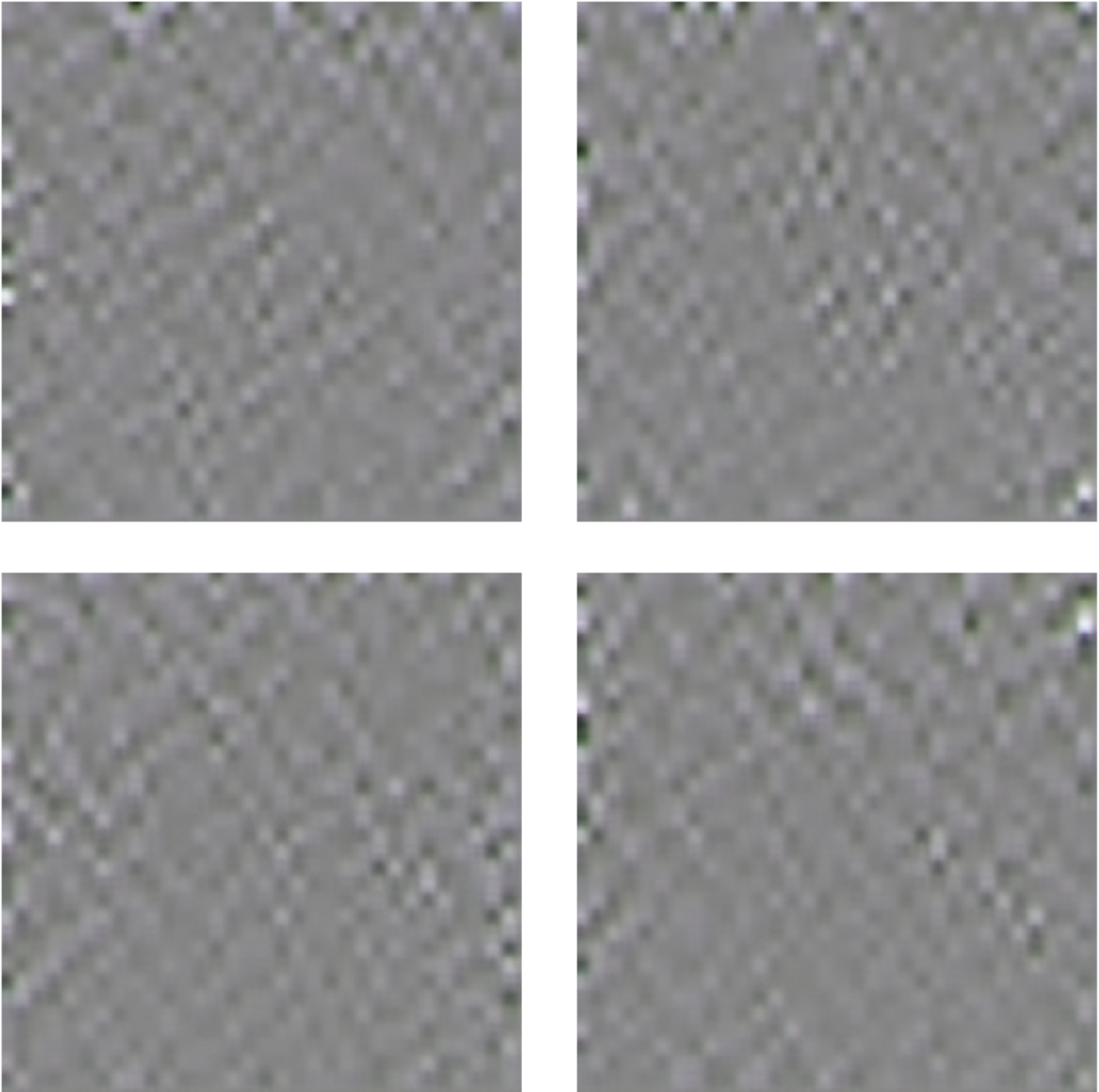} & 
    \includegraphics[width=0.3\textwidth]{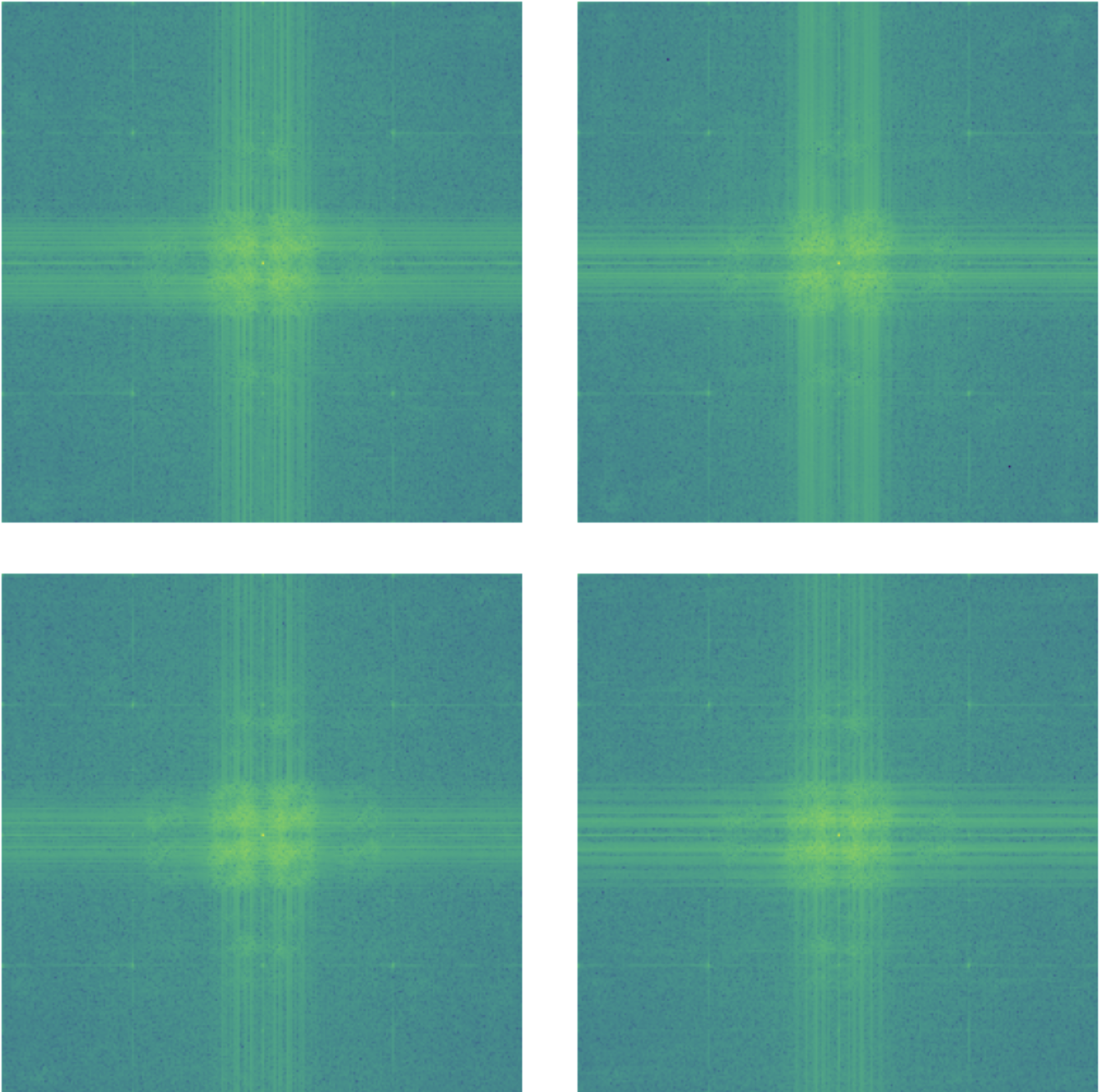} & 
    \includegraphics[width=0.35\textwidth]{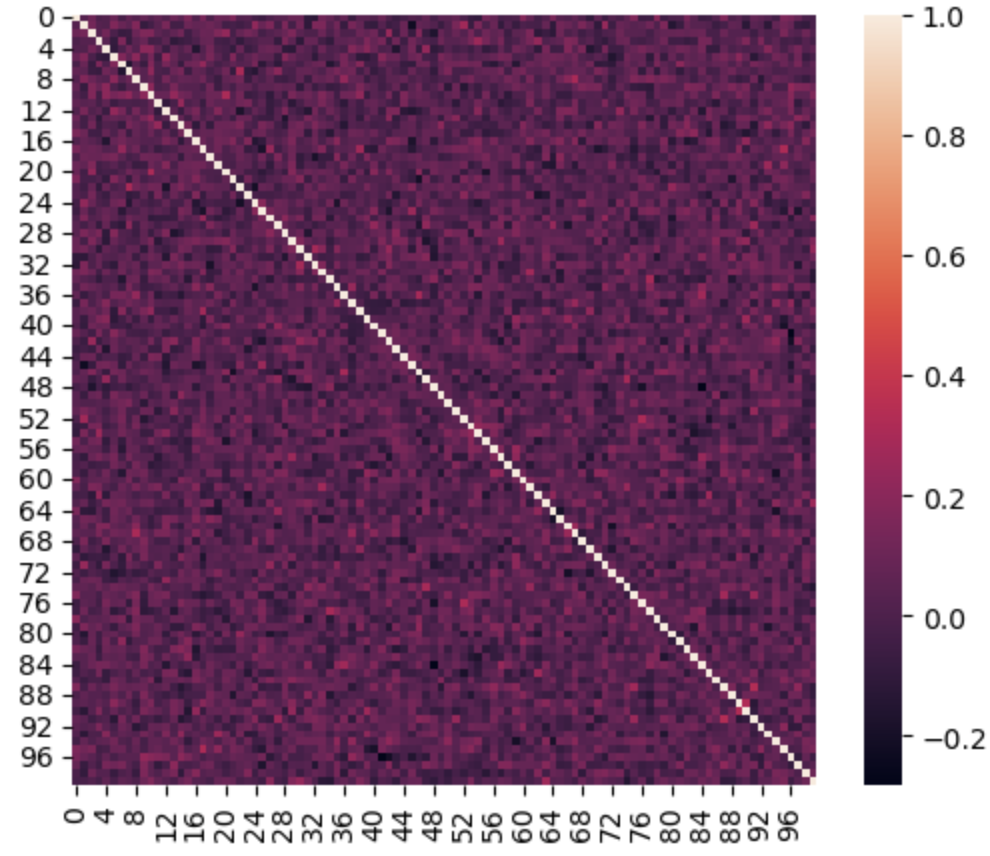}
\end{tabular}
\caption{Watermark Visualization: Spatial domain, Fourier domain and inter-watermark cosine similarity for $100$ watermarks. }
\label{fig:supp_watermark}
\end{figure*}

We visualize some sample watermarks in both, spatial and frequency domain in~\cref{fig:supp_watermark}. These watermarks are converted from bit-sequences to spatial domain as described in Sec.~$3.4$. Visually, the watermarks appear indistinguishable from one another in both domains. Yet, their orthogonality is clearly demonstrated through the cosine similarity matrix, which we used to analyze $100$ different watermarks. This matrix reveals that the inter-watermark cosine similarity is consistently close to zero, decisively indicating the orthogonal nature of these watermarks.

\section{Implementation Details}
We train \methodName with LDM for $15K$ iterations with a batch size of $32$, using $8$ NVIDIA A$100$ GPUs for each experiment. 
We use the default parameters for optimizers as used in the official repository of~\cite{rombach2022high}.
The learning rate is set at $3.2e^{-5}$ for training LDM.

\begin{figure*}[t!]
\centering
\includegraphics[trim={0 -4 0 0},clip,width=1\textwidth]{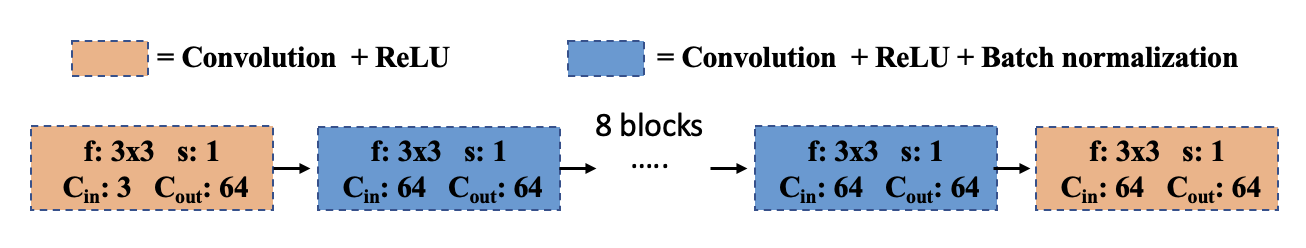}
\vspace{-8mm}
\caption{Generic decoder architecture. }
\label{fig:arch}
\vspace{-2mm}
\end{figure*}

We further show the architecture for the generic decoder used for comparing against pretrained secret decoder shown in~\cref{fig:arch}. The generic decoder consists of $2$ stem convolution layers and $10$ convolution blocks. Each block consists of convolutional and batch normalization layers followed by ReLU activation.

\section{More Sampled Images}
\begin{figure*}[t!]
\centering
\includegraphics[trim={0 -4 0 0},clip,width=1\textwidth]{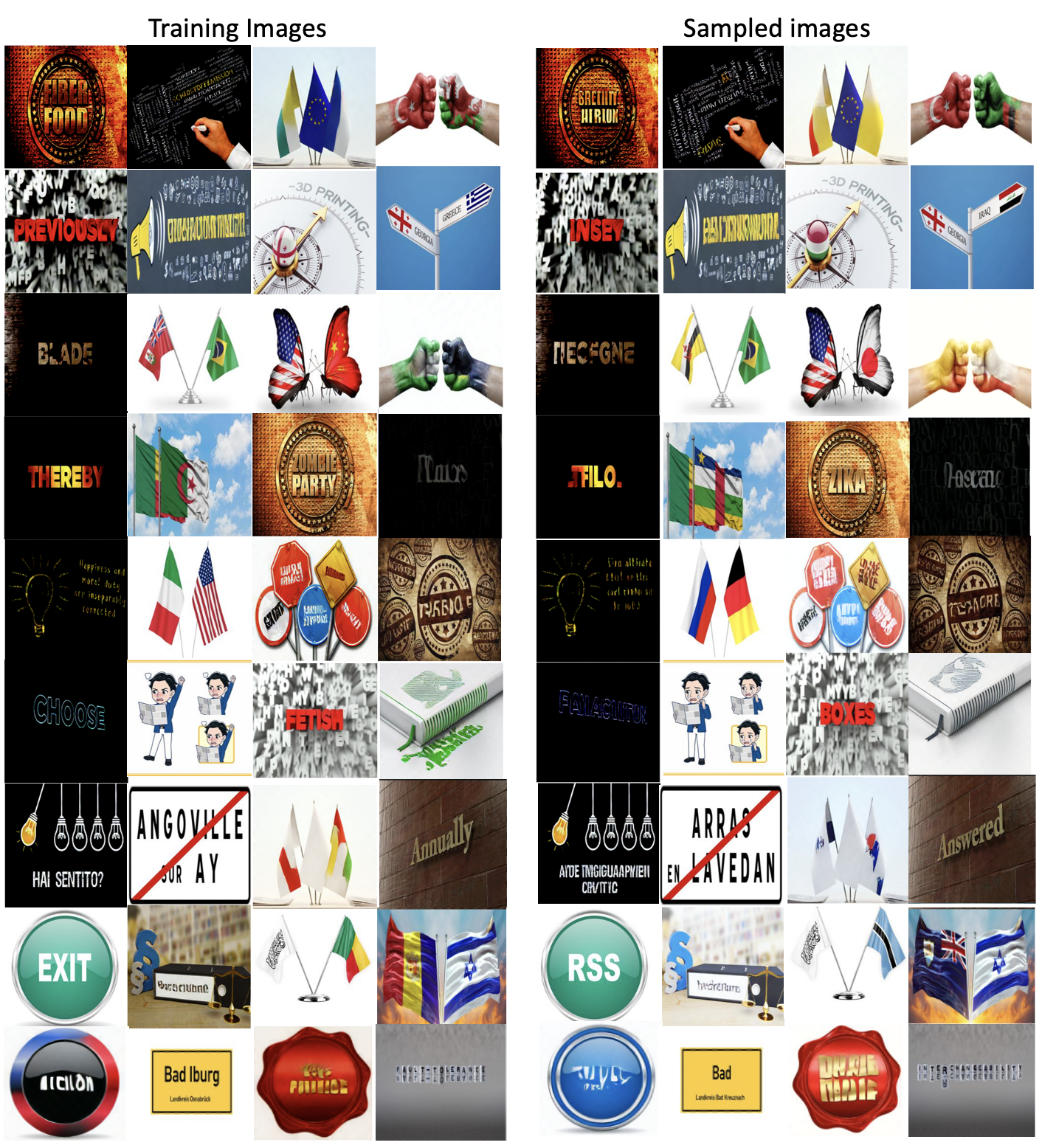}
\vspace{-4mm}
\caption{Training and sampled images for stock dataset. }
\label{fig:stock}
\vspace{-2mm}
\end{figure*}

\begin{figure*}[t!]
\centering
\includegraphics[trim={0 -4 0 0},clip,width=0.9\textwidth]{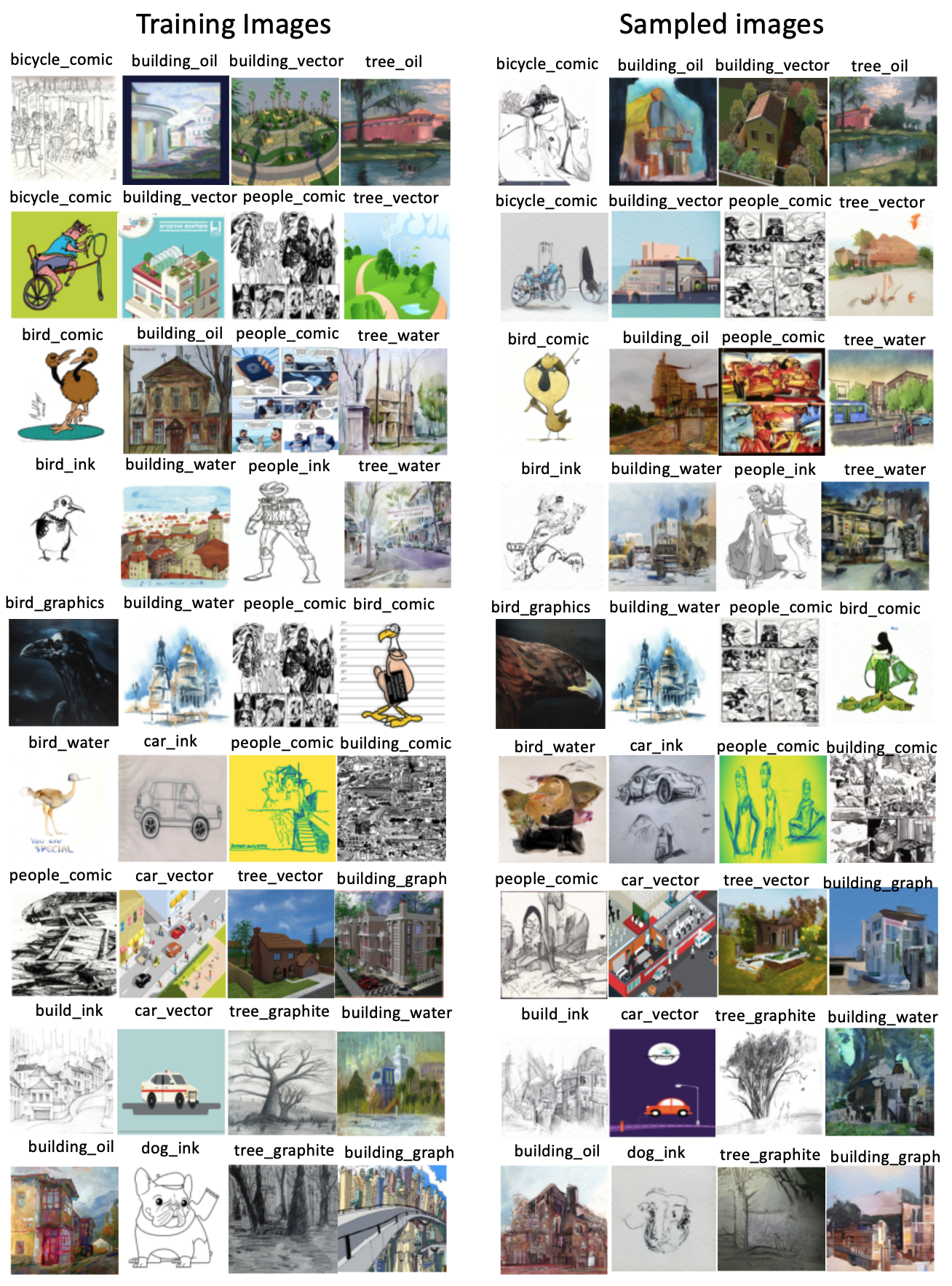}
\vspace{-4mm}
\caption{Training and sampled images for BAM dataset. }
\label{fig:bam}
\vspace{-2mm}
\end{figure*}

\begin{figure*}[t!]
\centering
\includegraphics[trim={0 -4 0 0},clip,width=1\textwidth]{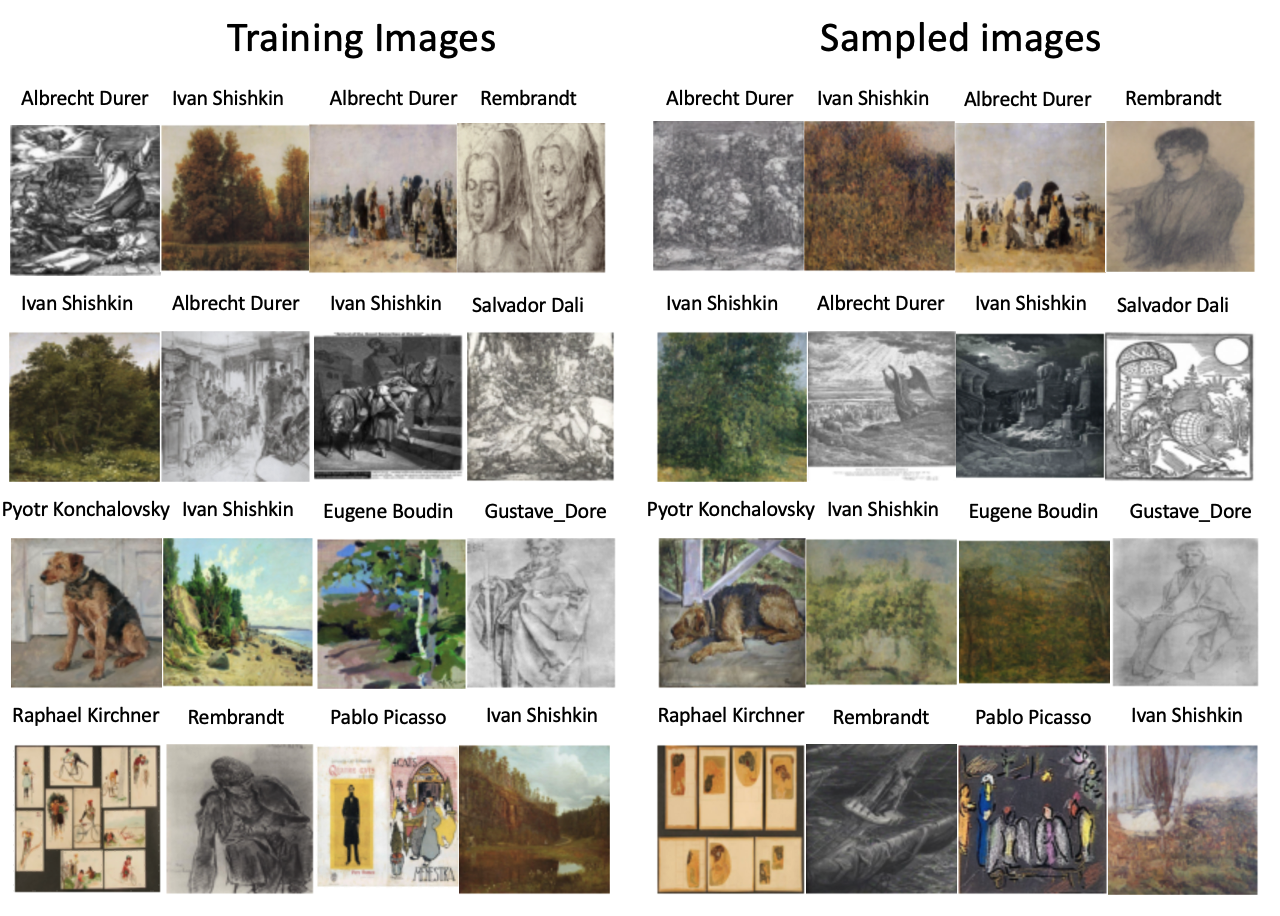}
\vspace{-8mm}
\caption{Training and sampled images for wiki-a dataset. }
\label{fig:wiki_a}
\vspace{-2mm}
\end{figure*}

\begin{figure*}[t!]
\centering
\includegraphics[trim={0 -4 0 0},clip,width=1\textwidth]{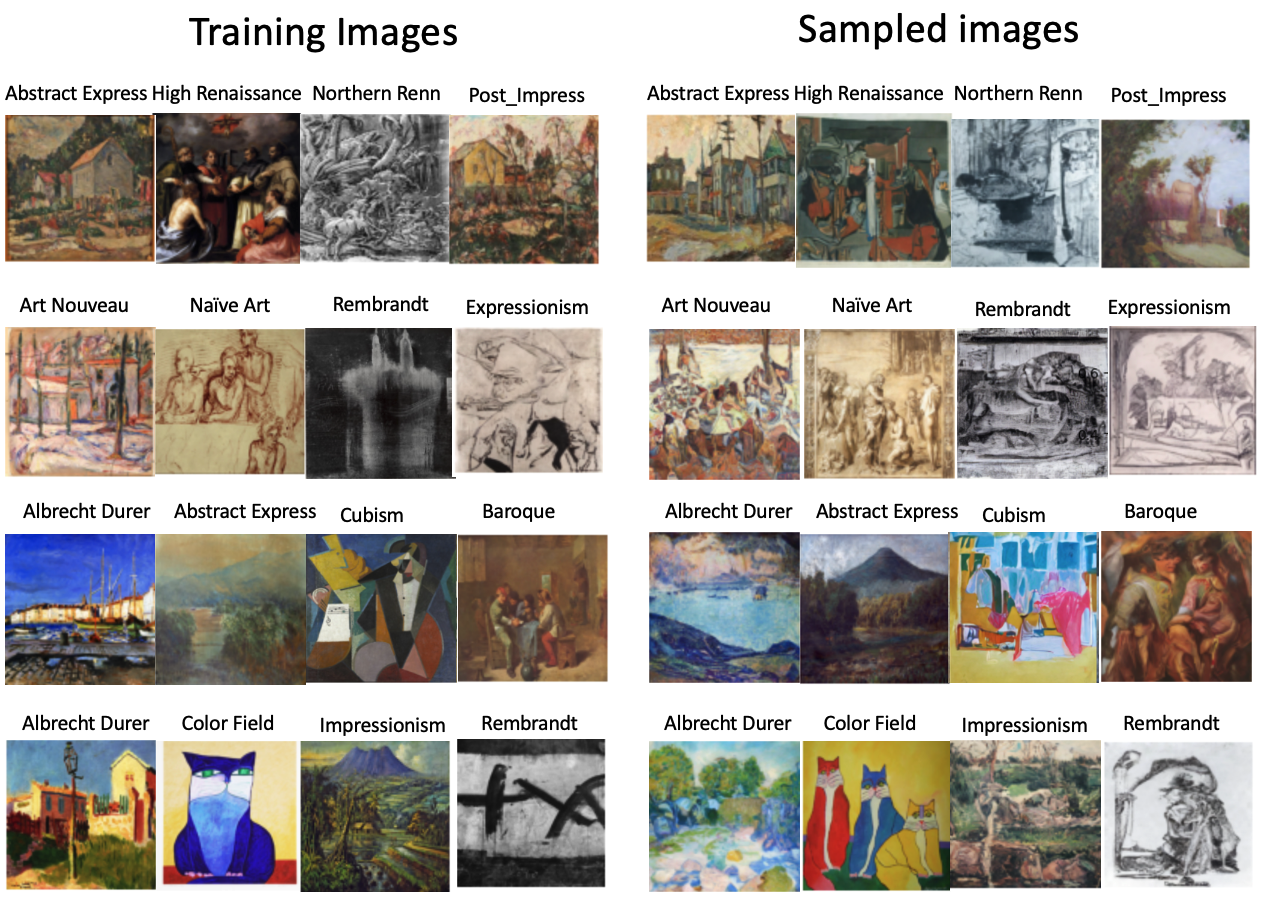}
\vspace{-8mm}
\caption{Training and sampled images for wiki-s dataset. }
\label{fig:wiki_s}
\vspace{-2mm}
\end{figure*}

We use multiple datasets for evaluating \methodName. We sample images from the trained LDM for every class. We show some of the train and sampled images for the corresponding classes for different datasets in~\cref{,fig:bam,fig:stock,fig:wiki_a,fig:wiki_s}. We argue that \methodName is able to perform attribution to different types of concepts, \ie image templates (\cref{fig:stock}), image style (\cref{fig:wiki_s}), style and content (\cref{fig:bam}), and ownership (\cref{fig:wiki_a}). Therefore, proactive based causal methods perform attribution not only on the style or motif of the image as done by correlation based works, but also performs attribution to a variety of concepts proving it's generalizability. 

\end{document}